\documentclass[]{beingbeyond}
\usepackage{enumitem}
\usepackage[toc,page,header]{appendix}

\usepackage[utf8]{inputenc} 
\usepackage[T1]{fontenc}    
\usepackage{hyperref}       
\usepackage{url}            
\usepackage{array}          
\usepackage{booktabs}       
\usepackage{amsfonts}       
\usepackage{nicefrac}       
\usepackage{microtype}      
\usepackage{xcolor}         
\usepackage{xspace}
\usepackage{bm}
\usepackage{bbm}
\usepackage{bbding}
\usepackage{tabularx}
\usepackage{textcomp}
\usepackage{amssymb}
\usepackage{enumitem}
\usepackage{amsmath}
\usepackage{mathtools}
\usepackage{amsthm}
\usepackage{multirow}
\usepackage{makecell}
\usepackage{color}
\usepackage{colortbl}
\usepackage{adjustbox}
\usepackage{caption}
\usepackage{graphicx}
\usepackage{wrapfig}
\usepackage{array}
\usepackage{multicol}
\usepackage{algorithm}
\usepackage{algorithmic}
\usepackage{diagbox}
\usepackage{cleveref}

\definecolor{myyellow}{RGB}{255,192,0}
\definecolor{mygreen}{RGB}{107,170,64}
\definecolor{mywrite}{RGB}{255,227,132}

\title{DemoHLM: From One Demonstration to Generalizable Humanoid Loco-Manipulation
}

\author{{\bfseries 
Yuhui Fu$^{1,2,*}$ \quad 
Feiyang Xie$^{1,2,*}$ \quad
Chaoyi Xu$^{2}$ \quad
Jing Xiong$^{1}$ \\
Haoqi Yuan$^{1,2}$  \quad
Zongqing Lu$^{1,2,\dagger}$
}}

\affiliation{{$^{1}$Peking University \quad $^{2}$BeingBeyond}}

\webpage{\url{https://beingbeyond.github.io/DemoHLM/}}

\abstract{
Loco-manipulation is a fundamental challenge for humanoid robots to achieve versatile interactions in human environments. Although recent studies have made significant progress in humanoid whole-body control, loco-manipulation remains underexplored and often relies on hard-coded task definitions or costly real-world data collection, which limits autonomy and generalization.
We present \textbf{DemoHLM}, a framework for humanoid loco-manipulation that enables generalizable loco-manipulation on a real humanoid robot from a single demonstration in simulation. DemoHLM adopts a hierarchy that integrates a low-level universal whole-body controller with high-level manipulation policies for multiple tasks.
The whole-body controller maps whole-body motion commands to joint torques and provides omnidirectional mobility for the humanoid robot. The manipulation policies, learned in simulation via our data generation and imitation learning pipeline, command the whole-body controller with closed-loop visual feedback to execute challenging loco-manipulation tasks.
Experiments show a positive correlation between the amount of synthetic data and policy performance, underscoring the effectiveness of our data generation pipeline and the data efficiency of our approach. Real-world experiments on a Unitree G1 robot equipped with an RGB-D camera validate the sim-to-real transferability of DemoHLM, demonstrating robust performance under spatial variations across ten loco-manipulation tasks.
}

\checkdata[Date]{October 11, 2025}

\definecolor{BlockC}{gray}{0.98}  
\definecolor{BlockA}{RGB}{191,211,230}
\definecolor{BlockB}{RGB}{199,233,192}

\begin{document}

\maketitle

\begingroup
\renewcommand\thefootnote{\fnsymbol{footnote}} 
\setcounter{footnote}{0}
\footnotetext[1]{Equal contribution.}
\footnotetext[2]{Correspondence to Zongqing Lu $<$lu@beingbeyond.com$>$.}
\endgroup

\section{Introduction}
Humanoid robots have become a central research focus in robotics due to their flexibility and adaptability to human environments. Advances in algorithms and hardware design have enabled expressive whole-body motions such as dance and kung-fu~\citep{asap, hover, kungfubot, gmt}. However, prior research has largely focused on learning whole-body controllers and humanoid teleoperation~\citep{openwbt, twist, clone, homie}, whereas developing autonomous policies for loco-manipulation remains underexplored.

Loco-manipulation requires contact-rich object interaction, coordinated whole-body joint control to reach arbitrary locations, and integration of visual inputs, posing significant challenges.
Prior work~\cite{dao2024sim,opt2skill,zhang2024wococo} studies reinforcement learning (RL) for loco-manipulation in simulation and demonstrates sim-to-real transfer on box-carrying tasks; however, these approaches rely on task-specific designs such as predefined subtask decompositions and task-specific reward functions, limiting their scalability to new tasks.
Recent studies~\citep{coohoi,liu2024mimicking,intermimic,pan2025tokenhsi} achieve scalable learning of humanoid–object interaction using RL and human–object–interaction datasets; however, their reliance on the SMPL~\citep{loper2023smpl} body model and special simulation parameters (e.g., large friction) hinders transfer to real humanoid robots.
Other work in whole-body control~\citep{he2024omnih2o,fu2024humanplus,homie,li2025amo} learns autonomous loco-manipulation policies from real-robot data via whole-body teleoperation and imitation learning, effectively reproducing human-like behavior to solve real-world tasks; nevertheless, scaling such costly real-world datasets to achieve spatial generalization and multi-task learning remains challenging.

In this study, we propose \textbf{DemoHLM}, a framework for humanoid loco-manipulation that enables scalable data generation and policy learning across diverse tasks. For each task, DemoHLM requires only a single demonstration and automatically synthesizes hundreds to thousands of successful trajectories that complete the same task in varied environments.
This approach originates from the MimicGen line of work~\citep{mimicgen, dexmimicgen, demogen}, where a single manipulation trajectory on fixed-base robot arms is used to synthesize additional trajectories under varied initial object poses via object-centric action replay. Extending this to humanoid loco-manipulation is non-trivial: the robot must coordinate whole-body joints to replay the end-effector trajectory while maintaining balance.

To address this challenge, we adopt a hierarchy that couples a low-level whole-body controller with high-level manipulation policies. The whole-body controller, trained via RL, controls the whole-body joint torques to track high-level commands, which include upper-body joint targets and torso motions. Building on this foundation, the manipulation policies operate in the high-level command space to learn complex humanoid–object interaction tasks.
In our data generation pipeline, we first collect a single successful demonstration in simulation via teleoperation, where human poses captured by a VR device are sent to the whole-body controller. The demonstration is segmented into three stages: locomotion, pre-manipulation, and manipulation. Given arbitrary initial objects and humanoid poses in simulation, the high-level commands in each stage are modified accordingly, concatenated into an action sequence, and replayed to generate a new trajectory. Finally, a behavior cloning policy is trained on all successful synthetic trajectories to serve as an autonomous loco-manipulation policy.
This pipeline enables scalable learning of generalizable manipulation behaviors without extensive teleoperation data collection, and the learned policy transfers zero-shot to a real humanoid robot.

We evaluate {DemoHLM} in both simulation and the real world, demonstrating its ability to accomplish a diverse suite of loco-manipulation tasks. Experiments show that increasing the amount of simulated data leads to substantial performance gains and improved generalization across varying initial states. Moreover, the generated data remain consistently effective across different behavior cloning algorithms, suggesting that the framework scalably produces high-quality data for different policy learning methods. On a real Unitree G1 humanoid robot, the learned policies achieve performance comparable to simulation across all tasks, demonstrating strong sim-to-real transferability.

Our contributions are summarized as follows:
\begin{itemize}
    \item We propose a simulation-based data generation pipeline for humanoid robots that can synthesize diverse successful trajectories from a single demonstration, enabling effective generalization of imitation learning policies for loco-manipulation.
    \item We introduce techniques that integrate a whole-body controller with object-centric motion planning to enable high-quality generation of humanoid loco-manipulation data.
    \item We evaluate the framework across a broad suite of tasks both in simulation and on a real humanoid robot, demonstrating its robustness and data efficiency in learning humanoid loco-manipulation and performing sim-to-real deployment. 
\end{itemize}

\section{Related Work}
\noindent\textbf{Humanoid Whole-Body Control}. 
In recent years, learning-based methods have significantly advanced the development of whole-body controllers for humanoid robots~\citep{li2025amo, openwbt, twist, gmt}. This progress has simplified the high-dimensional action space of humanoid robots and enabled more complex and refined loco-manipulation tasks. For example, TWIST~\citep{twist} proposes a unified keypoint-based controller trained via reinforcement learning and behavior cloning, achieving real-time teleoperation with rich whole-body skills.  AMO~\citep{li2025amo} integrates trajectory optimization with RL to develop an adaptive whole-body controller that follows torso velocity and rotation commands. OpenWBT~\citep{openwbt} organizes primitive skills into a structured skill space, facilitating robust planning and control. GMT~\citep{gmt} develops a tracking-based whole-body controller to track diverse motions in the real world by the motion Mixture-of-Experts architecture and an adaptive sampling strategy. To simplify the action space and accelerate learning, we employ a whole-body controller that accepts low-dimensional commands as the whole-body controller within our hierarchical architecture.

\noindent\textbf{Humanoid Loco-manipulation}.
Loco-manipulation couples whole-body balance with contact-rich manipulation under changing support conditions. Recent work advances this area by explicitly structuring contacts and skills across phases. For example, WoCoCo~\citep{zhang2024wococo} factorizes tasks into contact-phase primitives and learns compositional policies that generalize across diverse loco-manipulation behaviors. FALCON~\citep{zhang2025falcon} decouples locomotion and manipulation into coordinated agents with force-aware curricula, improving robustness to interaction forces and easing cross-robot transfer. Being-0~\citep{yuan2025being} integrates individual locomotion and manipulation skills into an agent framework to enable long-horizon navigation and manipulation tasks. 
Some recent works in whole-body control~\citep{he2024omnih2o, fu2024humanplus, li2025amo} further enable learning end-to-end loco-manipulation policies by collecting real-robot data using their whole-body teleoperation systems.
Other approaches~\citep{coohoi,liu2024mimicking,intermimic,pan2025tokenhsi} adopt RL for generalizable loco-manipulation using simulated SMPL humanoids, but have not demonstrated comparable success on real-world humanoid robots. 
In summary, existing methods either rely on task-specific design, require large amounts of real-world teleoperation data, or fail to deploy on real robots. 
In this work, we introduce a scalable approach that integrates simulation-based data generation with a whole-body controller, enabling the learning of generalizable humanoid loco-manipulation policies and effective sim-to-real transfer.

\noindent\textbf{Data Generation for Robotic Manipulation}.
Imitation learning methods such as Behavioral Cloning rely on large quantities of expert demonstrations to train visuomotor policies~\citep{pomerleau1988alvinn, 8461249, pmlr-v15-ross11a, ho2016generative}. However, collecting such datasets via human teleoperation can be costly and time-consuming~\citep{brohan2022rt, Khazatsky-RSS-24}. To alleviate this bottleneck, researchers have explored replay-based methods and data augmentation to enhance sample efficiency~\citep{pitis2020counterfactual, di2022learning}. MimicGen~\citep{mimicgen} addresses this challenge by automatically generating diverse demonstrations from a small seed set, and has inspired follow-up efforts in generative augmentation~\citep{chen2023genaug}, semantically imagined experiences~\citep{yu2023scaling}, bimanual dexterous manipulation~\citep{dexmimicgen}, and reinforcement learning~\citep{zhou2024learning,yuan2025demograsp,zhu2025dexflywheel}. 
However, prior works focus on fixed-base manipulators, leaving the challenges of humanoid loco-manipulation unaddressed. We bridge this gap with a novel data generation pipeline that leverages a whole-body controller to synthesize training data for humanoid loco-manipulation tasks.

\section{Method}

\begin{figure}[!t]
    \centering
    \includegraphics[width=\linewidth, trim={0cm, 1.5cm, 0cm, 1cm}, clip]{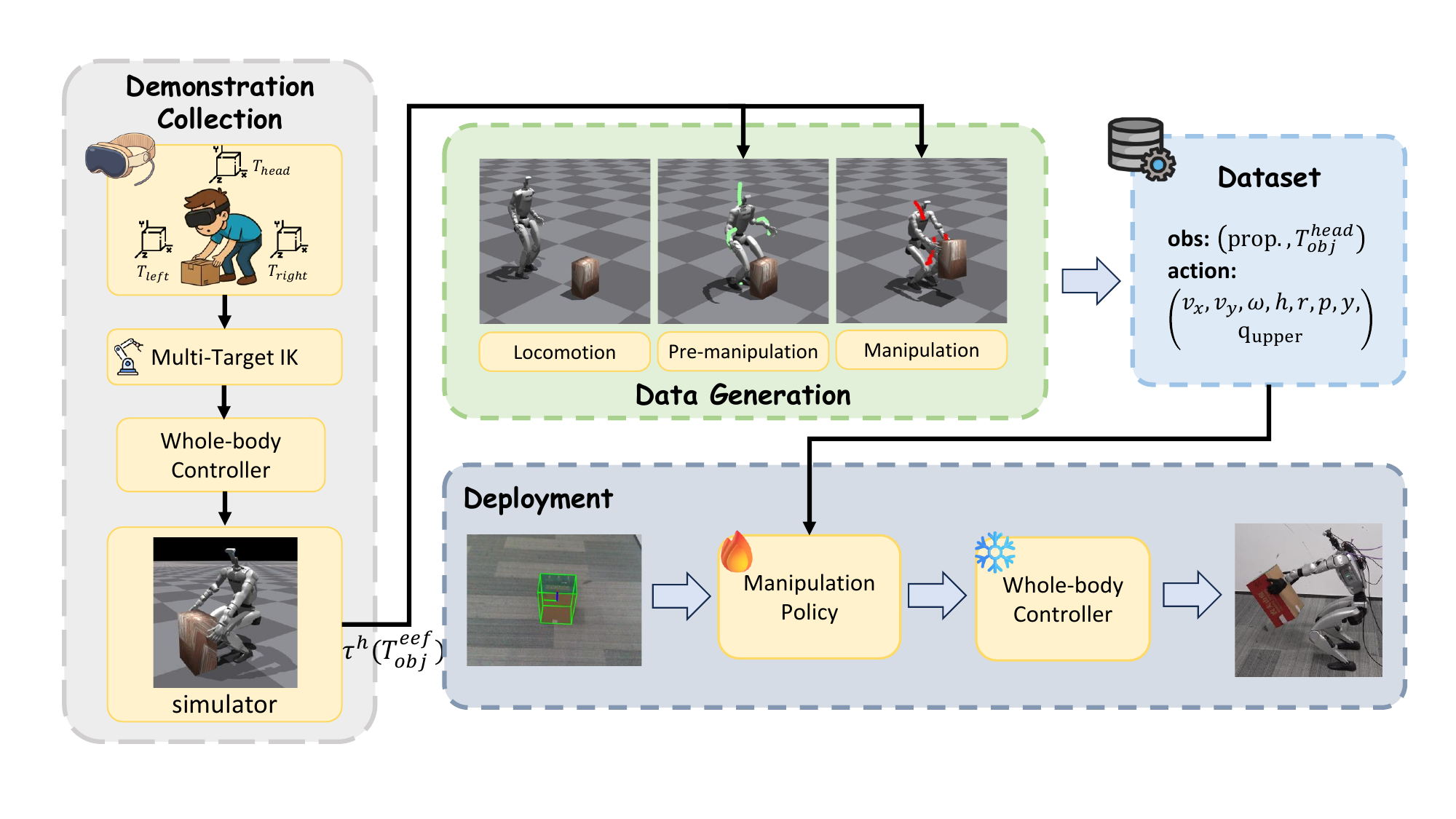}
    \caption{\textbf{Overview of {DemoHLM}.} For each task, we collect a single demonstration via VR teleoperation in simulation and record the robot trajectory in the object frame. This trajectory is then used to generate the pre-manipulation and manipulation phases in our data generation pipeline. The generated transitions include robot proprioception, object poses in the camera frame, and actions expressed as high-level commands sent to the whole-body controller. A manipulation policy is trained using imitation learning on this dataset and is successfully deployed on a real robot to perform loco-manipulation.}
    \label{fig:pipeline}
\end{figure}

We introduce \textbf{DemoHLM}, consisting of a simulation-based data-generation pipeline and a control hierarchy for humanoid loco-manipulation (Figure~\ref{fig:pipeline}). The framework enables the systematic collection of diverse training data while decomposing the control problem into two levels. Specifically, Sec.~\ref{method:data_gen} outlines the data-generation process, Sec.~\ref{method:low_level} details the RL-trained whole-body controller that provides robust whole-body control, and Sec.~\ref{method:high_level} presents the high-level manipulation policy responsible for task-directed decision making.

\subsection{Data Generation}
\label{method:data_gen}
To solve a task $\mathcal{M}$, we collect a single human manipulation demonstration $\tau^{h}=\{s^{h}_{t}\}_{t=1}^{T}$ in simulation, where $s_t^h$ contains the joint positions $\mathbf{q}_{\text{robot}}$, the poses of end effectors $\mathbf{p}^{\text{eef}}$ and objects $\mathbf{p}^{\text{obj}}$ in world frame. For notational simplicity, we omit the subscript $W$ (denoting the world frame) from poses expressed relative to the world frame. Our goal is to synthesize a large dataset $\mathcal{D}=\{\tau_{i}\}_{i=1}^{N}$ for $\mathcal{M}$ (with randomized initial robot and object poses) based on $\tau^{h}$. To achieve this, we first generate a dataset of the target poses of end effectors $D^{\text{eef}} = \{\tau_i^{\text{eef}}\}_{i=1}^{N}$ by converting $\tau^{h}$ into object-centric end effector pose sequences, where $\tau_i^{\text{eef}} = \{\hat{\mathbf{p}}^{\text{eef}}_t\}_{t=1}^T$. Then we input these target end-effector poses into our low-level controller to collect $\tau_{i}$ in simulation.

\subsubsection{Transforming the Human Demonstration into Object-Centric Poses}

At the first contact time $t_c$ in the human demonstration, we split $\tau^h$ into a pre-contact segment $\tau^o=\{s^h_t\}_{t=1}^{t_c}$ and a post-contact segment $\tau^p=\{s^h_t\}_{t=t_c}^{T}$. The transformation consists of the object-centric phase and the proprioception-centric phase. 
In object-centric phase, $\tau^o$ is transformed into end effector poses relative to the object’s pose. Specifically, The object-centric poses of the end effectors are defined as $T^{\text{eef}}_{\text{obj}} = T^{\text{eef}}(\bar{T}^{\text{obj}})^{-1}$, where $T^{A}_{B}$ denotes the pose of frame $A$ with respect to frame $B$ and $\bar{T}$ denotes the pose in task execution.
In proprioception-centric phase, $\tau^p$ is transformed into end effector poses relative to the agent’s own pose at timestep $t_c$. The proprioception-centric poses are defined as $T^{\text{eef}}_{\text{pro}} = T^{\text{eef}}(\bar{T}^{\text{eef}})_{t_c}^{-1}$. Finally, we concatenate the two type of poses to obtain the trajectory of relative poses $\tau^{\text{rel}}_i = \{ (T^{\text{eef}}_{\text{obj}})_1, \cdots, (T^{\text{eef}}_{\text{obj}})_{t_c}, (T^{\text{eef}}_{\text{pro}})_{t_c+1}, \cdots, (T^{\text{eef}}_{\text{pro}})_N \}$.

\subsubsection{Data Generation in Simulation}

For task $\mathcal{M}$, we generate the target end-effector poses trajectory $\tau_i^{\text{eef}}$ in the simulation through three stages: locomotion stage, pre-manipulation stage, and manipulation stage. We then input $\tau_i^{\text{eef}}$ into the low-level controller to collect trajectory $\tau_{i} = (s_1, a_1, \cdots, s_T)$ required for policy training. Here, $s_t = \{\mathbf{q}^{\text{robot}}_t, \dot{\mathbf{q}}^{\text{robot}}_t, r_t, p_t, \mathbf{p}^{\text{obj}}_t\}$ includes the body joint positions $\mathbf{q}^{\text{robot}}_t$, body joint velocities $\dot{\mathbf{q}}^{\text{robot}}_t$, roll and pitch angles $r_t$ and $p_t$, and the pose of the object in the camera frame $\mathbf{p}^{\text{obj}}_t$.

\paragraph{Locomotion Stage}
Since the human demonstration only includes the manipulation on objects, the robot must perform locomotion to reach a position near the starting position of $\tau^h$ when its initial position is far from the object. To guide the root of the robot to a suitable starting position, we employ a PD controller to generate velocity commands $(v_x, v_y, \omega_{yaw})$, which are then executed by the whole-body controller in the environment. The locomotion stage concludes when the distance and the relative rotation fall below a certain threshold.

\paragraph{Pre-Manipulation Stage}
This stage is based on the object-centric trajectory $\tau^o$. To ensure the end-effector trajectories produce similar effects under different initial conditions, our target end-effector poses are defined as: $\hat{T}^{\text{eef}} = T^{\text{eef}}_{\text{obj}} \bar{T}^{\text{obj}}$. Due to slight errors in the whole-body controller, the robot may not reach the exact starting point as in the human demonstration, resulting in a discrepancy between the current end-effector pose $\bar{T}^{\text{eef}}$ and the initial target pose $\hat{T}^{\text{eef}}_{\text{init}}$. To address this, we introduce an interpolated trajectory $\hat{\tau}^{\text{in}} = \{ (\hat{T}^{\text{eef}})^{\text{in}}_1, \cdots, (T^{\text{eef}})^{\text{in}}_K \}$ from $\bar{T}^{\text{eef}}$ to $\hat{T}^{\text{eef}}_{\text{init}}$ during the initial phase of the pre-manipulation stage. Consequently, the complete target trajectory for this stage becomes $\hat{\tau}^{o} = (\hat{\tau}^{in}, \{(\hat{T}^{\text{eef}})_i\}_{i=1}^{t_c})$.

\paragraph{Manipulation Stage}
In the manipulation stage, the end-effectors may remain largely stationary relative to the object in some common tasks (such as lifting a cube), which is hard to handle for generating data that only uses object-centric end-effector poses. To address this, we switch from object-centric poses to proprioception-centric poses just before manipulating with the object. This means the target end-effector poses in manipulation stage are given by $\hat{T}^{\text{eef}} = \bar{T}^{\text{eef}} (T^{\text{eef}}_{\text{pro}})^{-1}$, and the target trajectory is $\hat{\tau}^{p} = \{(\hat{T}^{\text{eef}})_i\}_{i=t_c}^{T}$.

Finally, we integrate the three stages and use the resulting trajectory as the target input to the whole-body controller to collect $\tau_i$ for training policies.

\subsection{Low-Level Whole-body Controller}
\label{method:low_level}
Several powerful whole-body controllers have been proposed~\citep{openwbt, clone}. Following AMO~\citep{li2025amo}, we adopt its controller as our low-level policy. The controller takes as input a set of desired motion parameters
\[
(v_x, v_y, \omega, h, r, p, y, \mathbf{q}_{\text{upper}}),
\]
where $v_x$ and $v_y$ are the linear velocities along the sagittal and lateral axes, $\omega$ is the yaw rate, $(h,r,p,y)$ specifies the desired torso configuration (height, roll, pitch, yaw), and $\mathbf{q}_{\text{upper}}$ are the desired upper-body joint positions. The controller outputs target joint positions for the full robot, $\mathbf{q}_{\text{target}}$, which are tracked via PD control.

Leveraging this whole-body policy markedly reduces the effective action dimensionality for the high-level manipulation policy, simplifying learning. Consistent with the original setup, the controller runs at $50\,\mathrm{Hz}$. This whole-body controller maintains balance and exhibits reliable sim-to-real transfer.


For the 2-DoF head camera, the objective is to keep the target near the image center. We use a proportional controller that regulates the neck joints using the target’s position in the camera frame, $\mathbf{p}_c=(x_c,y_c,z_c)^\top$. The commanded angular velocities are
\begin{equation*}
\omega_{\text{yaw}}   = \operatorname{clip}\!\big(-k_{\text{yaw}}\,y_c,\,-\omega_{\max},\,\omega_{\max}\big),\qquad
\omega_{\text{pitch}} = \operatorname{clip}\!\big(-k_{\text{pitch}}\,x_c,\,-\omega_{\max},\,\omega_{\max}\big),
\end{equation*}
where $k_{\text{yaw}},k_{\text{pitch}}>0$ are scalar gains and $\operatorname{clip}(u,a,b)=\min(\max(u,a),b)$.

At timestep $t$, joint angles are updated by integrating the angular velocities:
\begin{align*}
\theta_{\text{yaw}}(t+\Delta t)   &= \theta_{\text{yaw}}(t)   + \omega_{\text{yaw}}(t)\,\Delta t,\\
\theta_{\text{pitch}}(t+\Delta t) &= \theta_{\text{pitch}}(t) + \omega_{\text{pitch}}(t)\,\Delta t.
\end{align*}
This controller steers the camera to keep the target near the image center, mimicking human gaze stabilization during manipulation.

\subsection{High-Level Manipulation Policy}
\label{method:high_level}


For the manipulation policy, the objective is to learn task-relevant manipulation behaviors. The policy receives an observation vector
\[
\big[\;\mathbf{q}_{\text{pos}},\ \mathbf{q}_{\text{vel}},\ r,\ p,\ \mathbf{p}_{\text{obj}}^{\text{camera}}\;\big],
\]
where $\mathbf{q}_{\text{pos}}$ and $\mathbf{q}_{\text{vel}}$ are joint positions and velocities, $r$ and $p$ are the torso roll and pitch, and $\mathbf{p}_{\text{obj}}^{\text{camera}}$ is the estimated 6D pose of the target object in the camera frame. The policy outputs the high-level command to the whole-body controller described above, thereby bridging task-level decision making and whole-body control.

To instantiate this high-level module, we evaluate several behavior-cloning (BC) architectures, including an MLP with action chunking, ACT~\citep{act}, and Diffusion Policy~\citep{diffusion_policy}. The manipulation policy runs at $10\,\text{Hz}$. And it is slower than the $50\,\text{Hz}$ whole-body controller, reflecting its need for temporally aggregated observations and longer-horizon reasoning, whereas the low-level controller emphasizes fast stabilization and tracking.

\section{Experiments}
\subsection{System Design}
\label{system}

To realize \textbf{DemoHLM}, we develop a large suite of simulation environments tailored to loco-manipulation tasks and a unified teleoperation system for collecting human demonstrations directly in simulation. To facilitate seamless sim-to-real transfer, we mirror the entire setup in the real world, as shown in Figure~\ref{system}.

\begin{figure}[!t]
    \centering
    \includegraphics[width=0.6\linewidth]{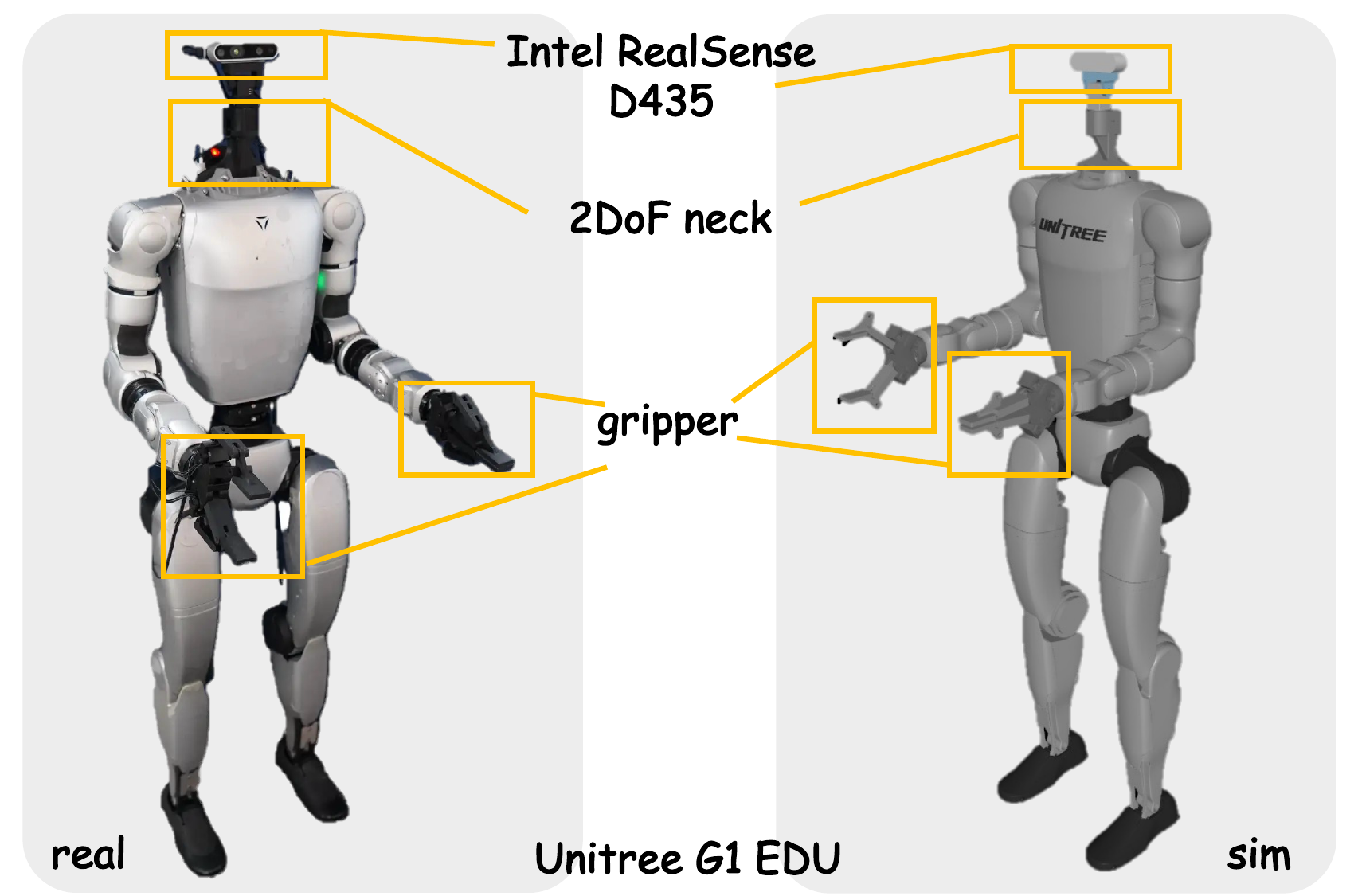}
    \caption{\textbf{Hardware Design.} We use a Unitree G1 humanoid robot in real-world experiments. To enable active vision, we mount a 2-DoF neck with an Intel RealSense D435 RGB-D camera. For tasks involving small objects, we attach parallel grippers to both end effectors.}
    \label{fig:system}
\end{figure}

\textbf{Simulation Environments.} All experiments are conducted in IsaacGym~\citep{isaacgym}. Task environments combine manually designed assets with objects rescaled from the PartNet-Mobility-v0 dataset in SAPIEN~\citep{Xiang_2020_SAPIEN,Mo_2019_CVPR,chang2015shapenet}. The robotic embodiment is a Unitree G1. To meet our method’s requirements, we augment the robot with a two-degree-of-freedom (2-DoF) active neck mounting an Intel RealSense D435 RGB-D camera. Because the stock G1 features compliant “rubber” hands that are insufficient for precise manipulation, we replace them with custom parallel-jaw grippers fabricated via 3D printing.

\textbf{Collecting Demonstrations.} We teleoperate in simulation to collect a single human demonstration per task. An Apple Vision Pro serves as the motion-capture device, using VisionProTeleop~\cite{visionproteleop} to stream head and wrist poses, which are mapped to the corresponding head and wrist links of the Unitree G1. The lower-body DoFs are fixed, and an additional prismatic joint along the $z$-axis provides vertical adjustment. We then solve a multi-task inverse-kinematics (IK) problem with \textsc{Pink}~\citep{pink} to obtain $(h,r,p,y,\mathbf{q}_{\text{upper}})$, which drives the robot to the desired end-effector poses.

\textbf{Real-World Setup.} We use the same robot and hardware modifications as in simulation. PD control is implemented via the Unitree SDK v2 (C++), wrapped with \texttt{pybind11}, and runs at $500\,\text{Hz}$. The Intel RealSense D435 is configured to $640{\times}480$ at $60\,\text{Hz}$, with depth frames aligned to the RGB stream. Dynamixel actuators drive both the neck and the grippers, enabling precise execution of commands. To ensure responsiveness, all I/O including image acquisition and motor encoder reads is handled asynchronously, preventing blockage of the main control thread and allowing the whole-body controller to operate reliably at $50\,\text{Hz}$.

\subsection{Task Settings}
\begin{figure}[!t]
    \centering
    \includegraphics[width=0.9\linewidth]{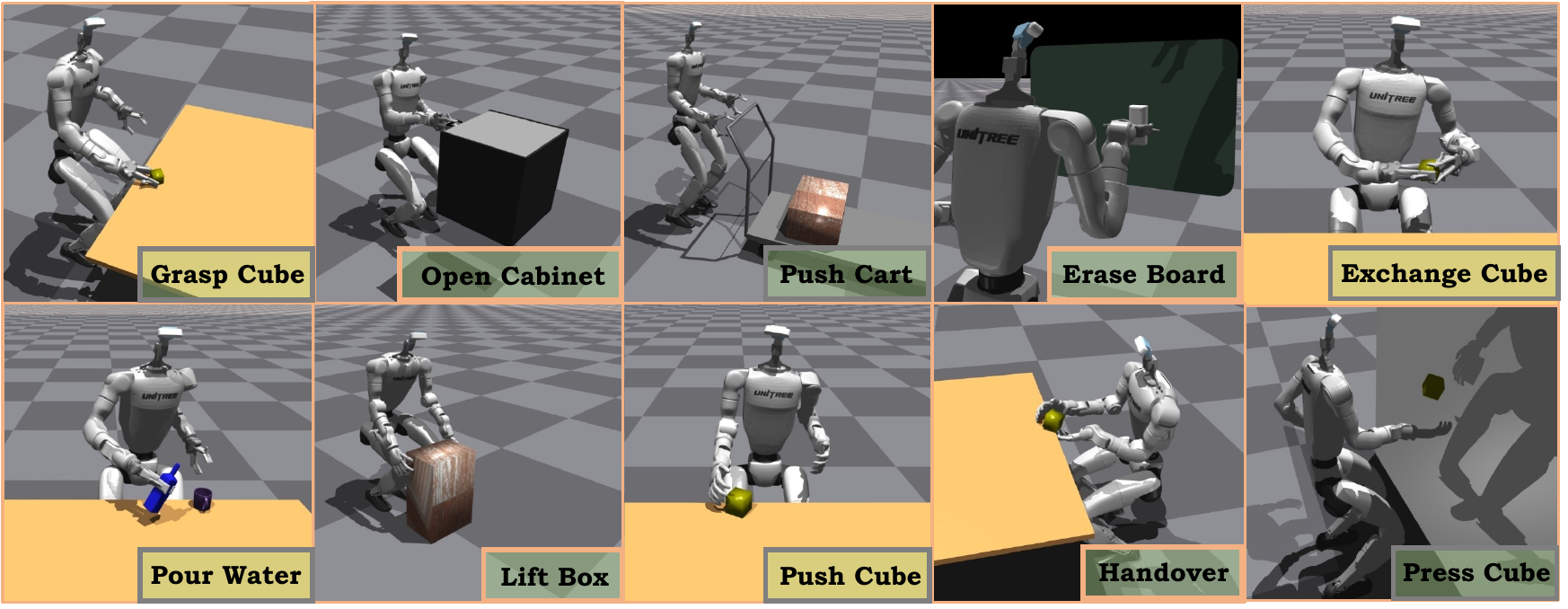}
    \caption{\textbf{Loco-manipulation Tasks.} We evaluate {DemoHLM} on ten tasks in both simulation and the real world. Four tasks can be completed using the rubber hands, while the remaining six tasks require parallel grippers for grasping and manipulation. Each task is initialized with randomized object and robot poses, requiring spatial generalization of the learned policies.}
    \label{fig:tasks}
\end{figure}

We design a suite of ten challenging loco-manipulation tasks shown in Figure \ref{fig:tasks} that require both approaching target objects and coordinating whole-body movements. These tasks are categorized into two groups: those that rely on rubber hands for interaction and those that require parallel grippers. 

\textbf{Rubber-hand tasks}
\begin{itemize}
    \item \textbf{LiftBox}: walk toward the box and lift it from the ground.  
    \item \textbf{PressCube}: walk toward a small cube and contact it.  
    \item \textbf{PushCube}: walk to a cube and push it along a specified direction.  
    \item \textbf{Handover}: manipulate a cube from the table using one hand and transfer it to the other.
\end{itemize}

\textbf{Gripper tasks}
\begin{itemize}
    \item \textbf{GraspCube}: approach a cube, grasp it with parallel grippers, and lift it.  
    \item \textbf{OpenCabinet}: reach and pull the handle of the cabinet using the grippers.  
    \item \textbf{PushCart}: walk toward a cart and push it.  
    \item \textbf{EraseBoard}: hold an eraser and wipe on a board surface.  
    \item \textbf{PourWater}: grasp a bottle and pour water into a container.  
    \item \textbf{ExchangeCube}: transfer a cube from one gripper to the other.  
\end{itemize}

\subsection{Simulation Experiments}
We conduct extensive experiments with DemoHLM and find that: (1) increasing the amount of synthetic data consistently improves task completion performance; (2) data generated by our pipeline benefits multiple BC methods, with method-dependent performance differences; and (3) as the initial-state distribution is widened (i.e., greater randomness in object and robot poses), the learned policies become more robust to state noise. We further evaluate success rates under varying initial states when generating data, validating the efficiency of DemoHLM.

\textbf{Policy success improves with more data, with diminishing marginal returns.} As shown in Table~\ref{tab:mainresult}, we train policies on datasets of varying sizes and evaluate each policy on 5{,}000 rollouts using three independent random seeds for statistical reliability. Across all tasks, success rates increase monotonically with dataset size; however, gains gradually taper off as the dataset grows, reflecting diminishing marginal returns.

\begin{table}[!t]
\centering
\caption{Success rates (\%) of all tasks in simulation given different sizes of generated datasets.}
\label{tab:mainresult}
\rowcolors{4}{gray!10}{white} 
\begin{tabularx}{1.0\linewidth}{
    >{\raggedright\arraybackslash}p{2.6cm} 
    *{5}{>{\centering\arraybackslash}X}
}
\toprule
\rowcolor{white}
 & \multicolumn{5}{c}{\textbf{Dataset Size}} \\
\cmidrule(lr){2-6}
\textbf{Task Name} & \textbf{100} & \textbf{200} & \textbf{500} & \textbf{1k} & \textbf{5k} \\
\midrule
LiftBox & $86.41^{\pm 0.20}$ & $94.06^{\pm0.23}$ & $96.20^{\pm0.54}$ & $98.60^{\pm0.10}$ & $98.83^{\pm0.19}$ \\
PressCube & $53.35^{\pm0.88}$ & $54.29^{\pm1.26}$ & $68.89^{\pm0.49}$ & $76.68^{\pm0.78}$ & $85.23^{\pm0.91}$  \\
PushCube & $52.39^{\pm0.55}$ & $63.07^{\pm0.51}$ & $68.49^{\pm0.48}$ & $73.15^{\pm0.72}$ & $89.25^{\pm0.06}$ \\
Handover & $28.54^{\pm0.54}$ & $38.99^{\pm0.74}$ & $46.95^{\pm1.02}$ & $52.13^{\pm0.88}$ & $57.50^{\pm0.42}$  \\
\midrule
GraspCube & $56.95^{\pm1.27}$ & $60.81^{\pm0.26}$ & $75.49^{\pm0.33}$ & $77.55^{\pm0.35}$& $87.86^{\pm0.73}$  \\
OpenCabinet & $18.86^{\pm0.23}$ & $42.48^{\pm0.57}$ & $49.51^{\pm1.26}$ & $54.91^{\pm0.56}$ & $67.30^{\pm0.56}$\\
PushCart & $75.81^{\pm0.18}$ & $85.33^{\pm0.53}$ & $92.63^{\pm0.19}$ & $93.41^{\pm0.36}$ & $95.79^{\pm0.05}$ \\
EraseBoard & $19.38^{\pm0.28}$ & $50.17^{\pm0.46}$ & $54.85^{\pm0.38}$ & $62.29^{\pm0.81}$ & $71.73^{\pm0.25}$ \\
PourWater & $19.81^{\pm0.18}$ & $26.38^{\pm0.14}$ & $36.52^{\pm0.42}$ & $44.53^{\pm0.23}$ & $58.72^{\pm0.20}$\\
ExchangeCube & $13.35^{\pm0.58}$ & $28.28^{\pm0.12}$ & $33.99^{\pm0.18}$ & $44.34^{\pm0.22}$ & $52.87^{\pm0.53}$\\
\bottomrule
\end{tabularx}
\par\medskip
\end{table}

\begin{table}[!t]
\centering
\caption{Success rates (\%) of different imitation learning methods.}
\label{tab:ablation_framework}
\rowcolors{2}{gray!10}{white} 
\begin{tabularx}{0.7\linewidth}{
    >{\raggedright\arraybackslash}p{2.4cm} 
    *{3}{>{\centering\arraybackslash}X}
}
\toprule
\rowcolor{white}
 & \multicolumn{3}{c}{\textbf{Network Framework}} \\
\cmidrule(lr){2-4}
\textbf{Task Name} & \textbf{ACT} & \textbf{MLP} & \textbf{DP} \\
\midrule
LiftBox & $98.83^{\pm0.19}$ & $85.60^{\pm0.73}$ & $96.97^{\pm0.54}$  \\
PressCube & $85.23^{\pm0.91}$ & $83.11^{\pm0.12}$ & $88.19^{\pm0.72}$ \\
PushCube & $89.25^{\pm0.06}$ & $71.54^{\pm0.84}$ & $90.19^{\pm0.49}$ \\
Handover & $57.50^{\pm0.42}$ & $36.47^{\pm0.43}$ & $54.87^{\pm0.65}$ \\
\midrule
GraspCube & $87.86^{\pm0.73}$ & $81.26^{\pm0.59}$ & $90.20^{\pm0.11}$ \\
OpenCabinet & $67.30^{\pm0.56}$ & $41.61^{\pm0.51}$ & $62.77^{\pm0.33}$ \\
PushCart & $95.79^{\pm0.05}$ & $90.20^{\pm0.19}$ & $96.11^{\pm0.22}$ \\
EraseBoard & $71.73^{\pm0.25}$ & $35.65^{\pm1.55}$ & $70.65^{\pm0.56}$ \\
PourWater & $58.72^{\pm0.20}$ & $31.22^{\pm0.14}$ & $58.70^{\pm0.38}$ \\
ExchangeCube & $52.87^{\pm0.53}$ & $15.05^{\pm0.55}$ & $53.55^{\pm0.26}$ \\
\bottomrule
\end{tabularx}
\par\medskip
\end{table}

\textbf{BC architectures differ in performance; more expressive models tend to perform better.}
We train policies with ACT~\citep{act}, an MLP, and Diffusion Policy~\citep{diffusion_policy} on the same dataset. For fairness, the MLP baseline uses action chunking, since both ACT and Diffusion Policy predict multi-step actions. As shown in Table~\ref{tab:ablation_framework}, ACT and Diffusion Policy achieve similar performance across tasks, whereas the simpler MLP exhibits a substantial drop in success rate. These results suggest that carefully designed, temporally expressive BC architectures yield stronger performance on loco-manipulation.

\textbf{Data-generation success decreases as the initial-state distribution broadens.}
Table~\ref{tab:ablation_region} reports the relationship between data-generation success and the size of the initial-state region for \emph{Handover} and \emph{GraspCube}. The decline is plausibly due to harder inverse-kinematics (IK) solves and increased likelihood of control-time anomalies (e.g., incidental collisions). Even for larger regions, \textsc{DemoHLM} adapts to randomized robot and object poses; although success rates drop modestly, the pipeline still yields sufficient high-quality trajectories for effective learning.

\begin{table}[!t]
\centering
\caption{Data-collection success rate (\%) in simulation with randomized initial states sampled from three regions, $R_1$–$R_3$, of increasing size. Exact parameters for each region are provided in Appendix~\ref{appendix:region}. Unless otherwise noted, we use $R_3$ as the default initial region in subsequent experiments.}
\label{tab:ablation_region}
\rowcolors{2}{gray!10}{white} 
\begin{tabularx}{0.5\linewidth}{
    >{\raggedright\arraybackslash}p{2.4cm} 
    *{3}{>{\centering\arraybackslash}X}
}
\toprule
\rowcolor{white}
 & \multicolumn{3}{c}{\textbf{Initial Region}} \\
\cmidrule(lr){2-4}
\textbf{Task Name} & \textbf{$R_1$} & \textbf{$R_2$} & \textbf{$R_3$} \\
\midrule
Handover & $0.997$ & $0.895$ & $0.792$  \\
GraspCube & $0.993$ & $0.971$ & $0.926$  \\
\bottomrule
\end{tabularx}
\par\medskip
\end{table}

\begin{figure}[!t]
    \centering
    \includegraphics[width=\linewidth]{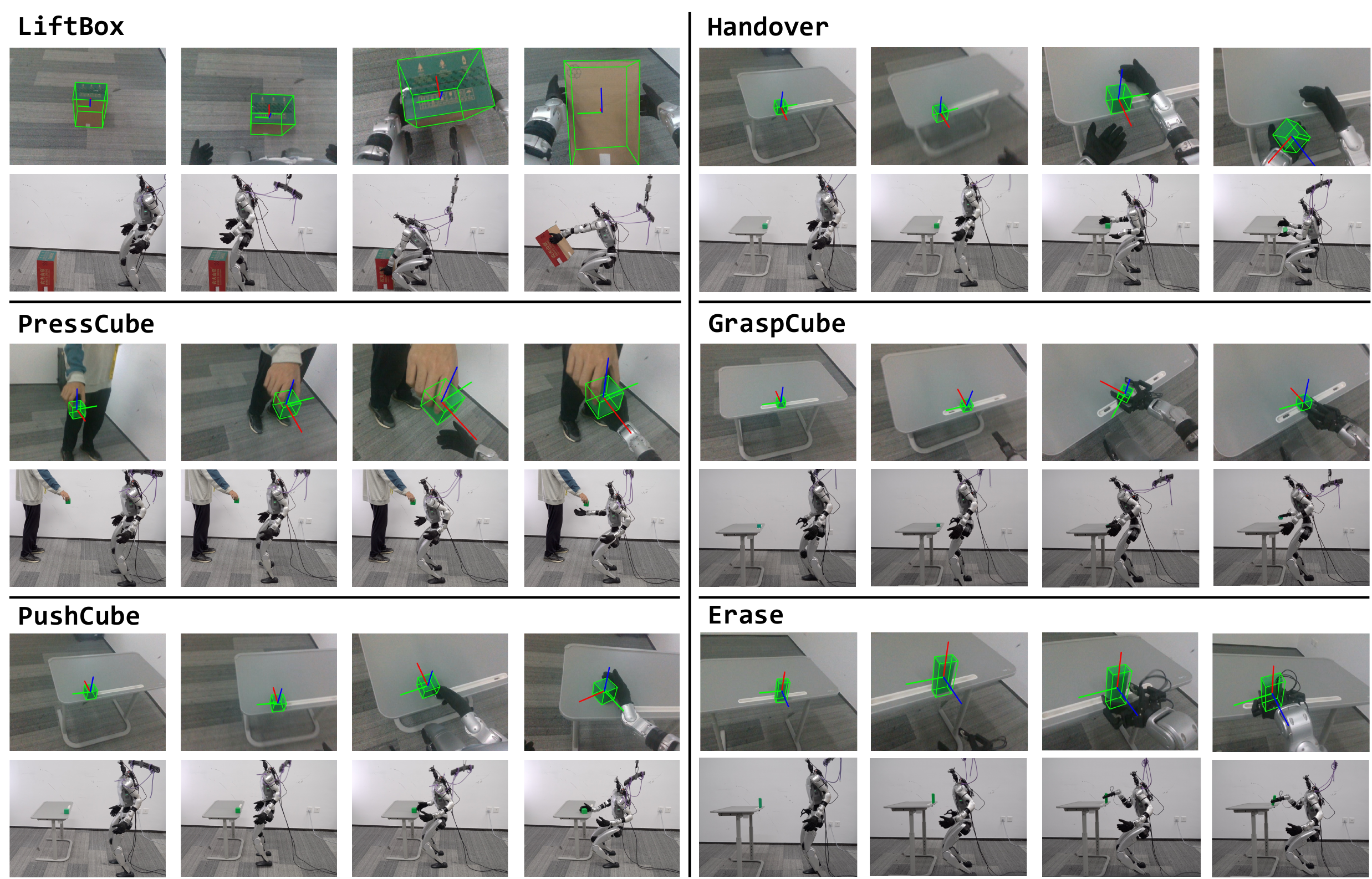}
    \caption{\textbf{Real-world policy rollouts.} Each pair of rows shows time-aligned first-person and third-person views. Frames progress from left to right over time.}
    \label{fig:real_1_and_3_view}
\end{figure}

\subsection{Sim-to-Real Transfer}

We validate that the policy trained with DemoHLM can be zero-shot transferred to real-world tasks. To enable sim-to-real transfer, we modified the G1 robot to match the configuration used in the simulator. We evaluated the policy on six tasks implemented in the simulator, where it demonstrated promising performance. In the real-world deployment, we employed a single head-mounted RealSense D435 camera to estimate object poses using FoundationPose~\citep{foundationpose}, ensuring consistency with the object pose representation in the simulator. Due to camera motion and frequent object jitter in the captured images, we employ FoundationPose++~\citep{foundationpose_plus_plus}, a real-time 6D pose tracker for highly dynamic scenes, as an enhanced perception module.

\begin{figure}[!t]
    \centering \includegraphics[width=0.95\linewidth]{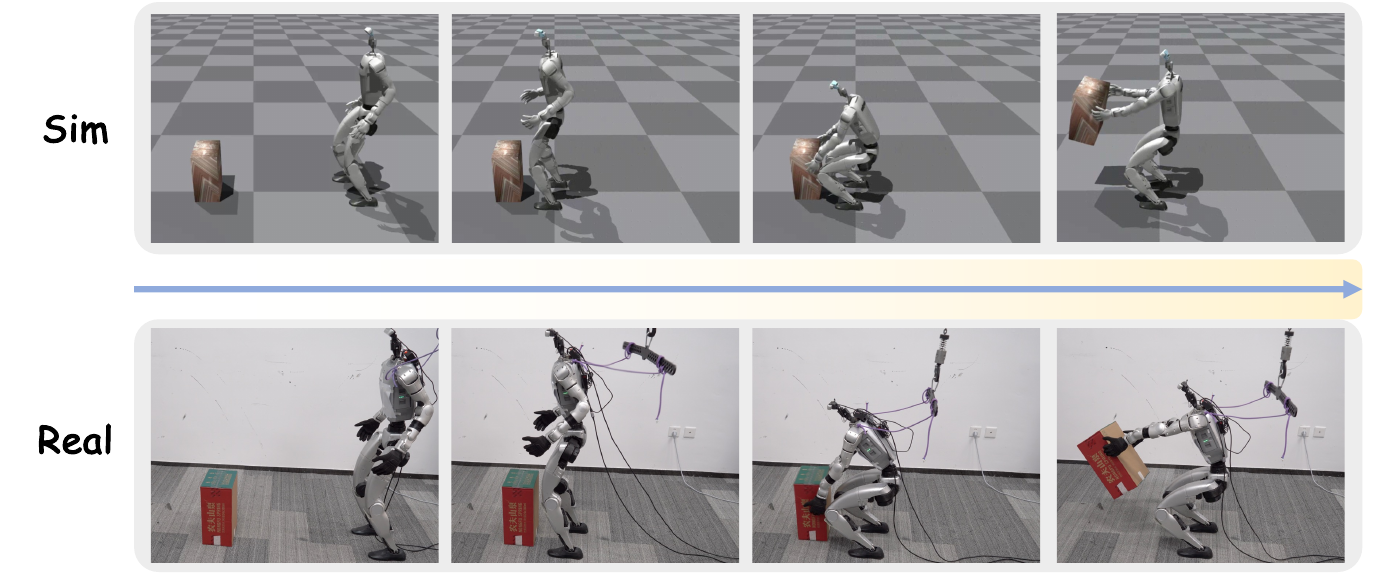}
    \caption{\textbf{Rollouts on {LiftBox} in simulation and the real world.} Frames are ordered left to right, illustrating key stages of the manipulation sequence executed by the learned policy.}
    \label{fig:sim-and-real}
\end{figure}

\begin{table}[!t]
\centering 
\caption{Success rates of real-world test.}
\label{tab:real}
\resizebox{\linewidth}{!}{
\begin{tabular}{cccccccc}
\toprule
Task Name & LiftBox & PressCube & PushCube & Handover & GraspCube & OpenCabinet & EraseBoard \\
\midrule
Success & $5/5$ & $5/5$ & $4/5$ & $4/5$ & $3/5$ & $2/5$ & $2/5$ \\
\bottomrule
\end{tabular}
}
\par\medskip
\end{table}

Figure~\ref{fig:real_1_and_3_view} presents real-world policy rollouts with both first-person views (with FoundationPose++ tracking results) and the corresponding third-person views. 
Table~\ref{tab:real} reports success rates for all tasks, demonstrating performance comparable to those in simulation.

Since the whole-body controller exhibits inconsistent tracking performance for velocity commands between simulation and reality, the tracking in the real world is less accurate. However, the high-level manipulation policy operates in a closed-loop manner, continuously adjusting velocity commands at a high frequency. This enables the humanoid robot to reach the desired positions, resulting in manipulation behaviors that remain largely consistent between simulation and reality, thereby allowing the completion of the tasks. Figure~\ref{fig:sim-and-real} compares simulation and real-world executions of {LiftBox} at the corresponding time steps.

\section{Conclusion and Limitations}

In this study, we propose {DemoHLM}, a data generation paradigm for learning humanoid loco-manipulation. DemoHLM synthesizes training data and learns loco-manipulation policies upon an RL-based whole-body controller. We show that this framework achieves robust and generalizable performance across ten challenging tasks in both simulation and the real world.

However, our current approach has several limitations. It relies exclusively on simulation-generated data, which can hinder the performance on the real robot due to the sim-to-real gap in environment dynamics and visual sensors. Using a single RGB-D camera may also constrain performance in cluttered or occluded scenes. In addition, using FoundationPose to provide the policy with object 6D poses limits the manipulation of unmodeled objects.
Future work could explore hybrid training on mixed simulation and real-robot data, including wrist cameras or tactile sensors, and visual sim-to-real approaches (e.g., using only RGB inputs) to address these limitations. 


\bibliography{iclr2026_conference}
\bibliographystyle{iclr2026_conference}

\newpage
\appendix

\section{Additional Results}

\begin{figure}[!h]
    \centering
    \includegraphics[width=0.9\linewidth]{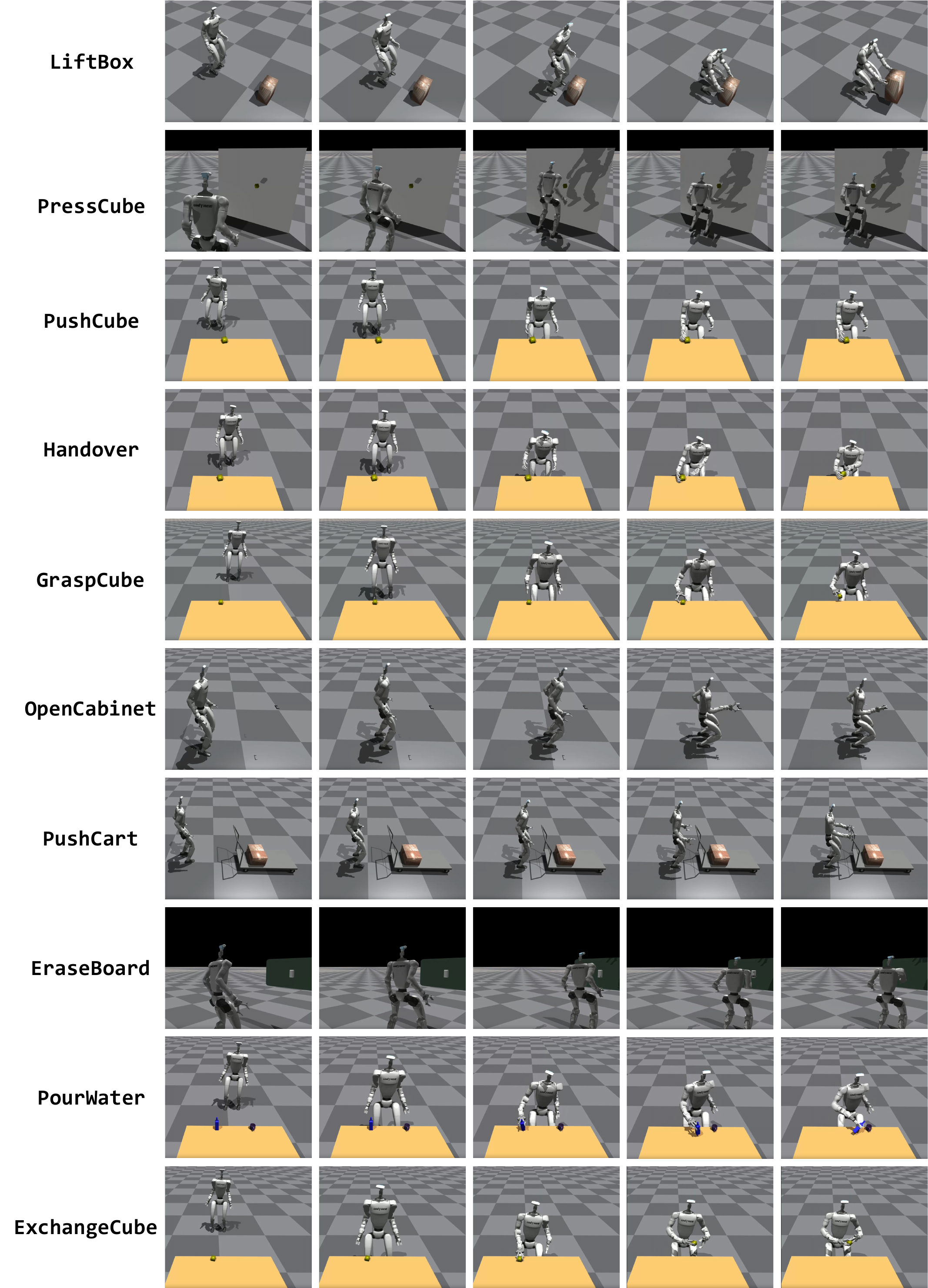}
    \caption{Policy rollouts in simulation, with images progressing from left to right over time.}
    \label{fig:sim_experiments}
\end{figure}

\subsection{Simulation Experiments}

Figure~\ref{fig:sim_experiments} shows the policy rollouts for all tasks in simulation.

\subsection{Data Generation Success Rate}
Due to occasional failures in inverse kinematics (IK) solving and unexpected events such as collisions during control, the data generation success rate varies across tasks. Table~\ref{tab:data_gen_sr} reports the success rates for different tasks under the maximal random initial state region.
\begin{table}[htbp]
\centering
\caption{Success rates (\%) of data generation for different tasks.}
\label{tab:data_gen_sr}
\rowcolors{2}{gray!10}{white} 
\begin{tabularx}{0.4\linewidth}{
    >{\raggedright\arraybackslash}p{2.4cm} 
    *{1}{>{\centering\arraybackslash}X}
}
\toprule
\rowcolor{white}
\textbf{Task Name} & \textbf{Data generation success rate}\\
\midrule
LiftBox & $91.2$ \\
PressCube & $92.5$  \\
PushCube & $99.3$ \\
Handover & $79.2$ \\
\midrule
GraspCube & $92.6$\\
OpenCabinet & $82.3$ \\
PushCart & $97.6$  \\
EraseBoard & $74.6$  \\
PourWater & $81.9$  \\
ExchangeCube & $72.7$ \\
\bottomrule
\end{tabularx}
\par\medskip
\end{table}

\section{Implementation Details}
\subsection{Hardware Design}

We employ the Unitree G1 EDU humanoid robot, which has 29 degrees of freedom (DoFs). To meet the requirements of our tasks—where the robot must continuously track and perceive target objects—we designed a custom 2-DoF neck equipped with an RGB-D RealSense D435 camera, replacing the original fixed head with a limited field of view. 

In addition, since the original robot is equipped only with two passive rubber hands without any actuation, we developed a pair of lightweight grippers to enable simple object manipulations and enrich the diversity of tasks. All custom hardware components were fabricated using 3D printing with high-strength PLA material to ensure both accessibility and reliability. Each joint of the modified structure is actuated using Dynamixel XC330-M288-T servo motors, which provide accurate and convenient control.

For control frequency settings, we adopt different rates for various layers of the system. In table-top manipulation tasks, a high control frequency leads to minimal observable state changes and repeated actions due to action chunking. Therefore, we set the high-level manipulation policy to operate at 10 Hz, the whole-body controller at 50 Hz, and the PD controller at 500 Hz to maintain balance and responsiveness.

To minimize inference latency, the entire system is deployed on an NVIDIA RTX 4090 GPU and communicates with the robot via Ethernet. All sensory observations are acquired asynchronously, as the data sources operate at different frequencies. The RGB-D camera streams 640 × 480 images at 60 Hz; the FoundationPose module asynchronously retrieves camera data at approximately 25 Hz; and the neck and grippers are connected via USB 3.0 with a baud rate of 1,000,000. Although asynchronous sensing leads to minor temporal misalignment compared to the simulator, the system remains stable and achieves successful task execution with negligible performance degradation.

\subsection{Task Settings}

\subsubsection{Initial Regions}
\label{appendix:region}
The initial regions refer to the range of randomization for the initial poses of robots and objects. For the robot, we randomize its initial x, y coordinates $x_{robot}, y_{robot}$ and yaw orientation $yaw_{robot}$. For the object, we randomize its initial y, z coordinates $y_{obj}, z_{obj}$ and z-direction rotation $z^{rot}_{obj}$ by add delta adding offsets $\Delta y_{obj}, \Delta z_{obj}, \Delta z^{rot}_{obj}$ on the default initial pose of each task. These variables are sampled from the corresponding uniform distributions in Table~\ref{tab:app-region}.

\begin{table}[htbp]
\centering
\caption{The exact parameters used for each initial region. }
\label{tab:app-region}
\rowcolors{2}{gray!10}{white} 
\begin{tabularx}{1.0\linewidth}{
    >{\centering\arraybackslash}p{1.2cm} 
    *{6}{>{\centering\arraybackslash}X}
}
\toprule
\rowcolor{white}
 & \multicolumn{6}{c}{\textbf{Exact Parameters}} \\
\cmidrule(lr){2-7}
\textbf{Regions} & \textbf{$x_{robot}$} & \textbf{$y_{robot}$} & \textbf{$yaw_{robot}$} & \textbf{$\Delta y_{obj}$} & \textbf{$\Delta z_{obj}$} & \textbf{$\Delta z^{rot}_{obj}$} \\
\midrule
$R_1$ & $(-0.6,-0.3)$ & $(-0.05,0.05)$ & $(-\frac{\pi}{16},\frac{\pi}{16})$ & $(0.0,0.02)$ & $(-0.05,0.0)$ & $(-\frac{\pi}{4},\frac{\pi}{4})$ \\
$R_2$ & $(-1.0,-0.3)$ & $(-0.1,0.1)$ & $(-\frac{\pi}{8},\frac{\pi}{8})$ & $(0.0,0.05)$ & $(-0.1,0.0)$ & $(-\frac{\pi}{2},\frac{\pi}{2})$ \\
$R_3$ & $(-1.2,-0.3)$ & $(-0.3,0.3)$ & $(-\frac{\pi}{4},\frac{\pi}{4})$ & $(0.0,0.1)$ & $(-0.2,0.0)$ & $(-\pi,\pi)$ \\
\bottomrule
\end{tabularx}
\par\medskip
\end{table}

\subsubsection{Success Conditions}
For each task, the success conditions are defined as follows.
\begin{itemize}
    \item \textbf{LiftBox}: The distance between the pelvis of robot and the center of box $d^{robot}_{obj} < 0.6m$ and the height of the center of box $h_{obj} > 0.6m$ (the initial height of the center of box is $0.29m$).  
    \item \textbf{PressCube}: The distance between the right hand of robot and the center of button $d^{righthand}_{obj} < 0.05m$.
    \item \textbf{PushCube}: The cube moves more than 0.1m in the positive y-axis direction.
    \item \textbf{Handover}: The distance between the left hand of robot and the center of button $d^{lefthand}_{obj} < 0.05m$, and the cube is lower than the table.
    \item \textbf{GraspCube}: The right gripper is in a closed state, the x-coordinate of the cube is less than $0.27m$, and the y-coordinate is $0.1m$ higher than the table.  
    \item \textbf{OpenCabinet}: The right gripper is in a closed state and the movement distance of handle exceeds $0.2m$.
    \item \textbf{PushCart}: Both left and right claws are in a closed state and gripping the crossbar.  
    \item \textbf{EraseBoard}: The right gripper grips the eraser and drives it to move more than $0.2m$ to the left.
    \item \textbf{PourWater}: The right gripper grips the water bottle and moves it to the left so that its distance from the center of the water bottle is less than $0.1m$.
    \item \textbf{ExchangeCube}: The left gripper clamps the cube and causes it to move in the y-direction by more than $0.3m$.
\end{itemize}

\subsection{Network Architectures and Training Hyperparameters}
We provide details of the network architectures and training hyperparameters for all considered policy models. Tables~\ref{tab:mlp_hyperparams} --~\ref{tab:dp_hyperparams} summarize the network architectures and training hyperparameters for the MLP, ACT, and Diffusion Policy models.
  
\begin{table}[H]
\centering 
\caption{The MLP architecture and training hyperparameters.}
\label{tab:mlp_hyperparams}
\begin{tabular}{cc}
\toprule
Hyperparameter& Value \\
\midrule
layer num & 4 \\
1-hidden dim & 512 \\
2-hidden dim & 2048 \\
learning rate & 5e-5 \\
batch size & 512 \\
chunk size & 20 \\
\bottomrule
\end{tabular}
\par\medskip
\end{table}

\begin{table}[H]
\centering 
\caption{The ACT architecture and training hyperparameters.}
\label{tab:act_hyperparams}
\begin{tabular}{cc}
\toprule
Hyperparameter& Value \\
\midrule
encoder layers & 2 \\
decoder layers & 3 \\
heads & 8 \\
hidden dim & 512 \\
feedforward dim & 3200 \\
latent dim & 32 \\
learning rate & 5e-5 \\
batch size & 2048 \\
kl weight & 10 \\
chunk size & 20 \\
\bottomrule
\end{tabular}
\par\medskip
\end{table}

\begin{table}[H]
\centering 
\caption{The Diffusion Policy architecture and training hyperparameters.}
\label{tab:dp_hyperparams}
\begin{tabular}{cc}
\toprule
Hyperparameter& Value \\
\midrule
Unet layers & 6 \\
step embedding dim & 256 \\
down dims & 256,512,1024 \\
kernel size & 5 \\
observation horizon & 1 \\
predict horizon & 20 \\
action horizon & 10 \\
batch size & 4096 \\
learning rate & 5e-5 \\
diffusion iters & 100 \\

\bottomrule
\end{tabular}
\par\medskip
\end{table}

\newpage
\section{Credit Authorship}

\begin{multicols}{2} 

\noindent\textbf{Initial Exploration}
\begin{itemize}
    \item \textsc{Yuhui Fu}
    \item \textsc{Feiyang Xie}
    \item \textsc{Haoqi Yuan}
    \item \textsc{Jing Xiong}
\end{itemize}

\noindent\textbf{Data Generation \& Training}
\begin{itemize}
    \item \textsc{Yuhui Fu}
    \item \textsc{Feiyang Xie}
\end{itemize}

\noindent\textbf{Sim-to-Real Pipeline}
\begin{itemize}
    \item \textsc{Yuhui Fu}
\end{itemize}

\columnbreak

\noindent\textbf{Hardware Design}
\begin{itemize}
    \item \textsc{Chaoyi Xu}
\end{itemize}

\noindent\textbf{Real-World Experiments}
\begin{itemize}
    \item \textsc{Yuhui Fu}
    \item \textsc{Feiyang Xie}
    \item \textsc{Jing Xiong}
\end{itemize}

\noindent\textbf{Project Lead}
\begin{itemize}
    \item \textsc{Haoqi Yuan}
    \item \textsc{Zongqing Lu}
\end{itemize}

\end{multicols} 





\clearpage

\end{document}